\pdfoutput=1

\documentclass[tinyml]{acmart}

\AtBeginDocument{%
  \providecommand\BibTeX{{%
    \normalfont B\kern-0.5em{\scshape i\kern-0.25em b}\kern-0.8em\TeX}}}

\setcopyright{rightsretained}
\copyrightyear{2025}
\acmYear{2025}

\usepackage{threeparttable}
\usepackage{pgfplots}
\pgfplotsset{compat=1.18}
\usepackage{tikz}
\usepackage{amsmath}
\usepackage{cases}
\usepackage{subfig}
\usepackage{xcolor,colortbl}
\usepackage{multirow}
\usepackage{url}
\usepackage{IEEEtrantools}

\usetikzlibrary{arrows.meta,positioning,automata}
\usetikzlibrary{backgrounds,positioning,shapes.gates.logic.US}
\usetikzlibrary{shapes.geometric}

\usepackage{collcell}
\usepackage{enumitem}

\begin{document}

\title{\textit{ETHEREAL}: Energy-efficient and High-throughput Inference using Compressed Tsetlin Machine}

\author{Shengyu Duan}
\affiliation{%
  \institution{Newcastle University}
  \city{Newcastle upon Tyne}
  \country{UK}}
\email{shengyu.duan@newcastle.ac.uk}

\author{Rishad Shafik}
\affiliation{%
  \institution{Newcastle University}
  \city{Newcastle upon Tyne}
  \country{UK}}
\email{rishad.shafik@newcastle.ac.uk}

\author{Alex Yakovlev}
\affiliation{%
  \institution{Newcastle University}
  \city{Newcastle upon Tyne}
  \country{UK}}
\email{alex.yakovlev@newcastle.ac.uk}


\begin{abstract}
The Tsetlin Machine (TM) is a novel alternative to deep neural networks (DNNs). Unlike DNNs, which rely on multi-path arithmetic operations, a TM learns propositional logic patterns from data literals using Tsetlin automata. This fundamental shift from arithmetic to logic underpinning makes TM suitable for empowering new applications with low-cost implementations. 

In TM, literals are often included by both positive and negative clauses within the same class, canceling out their impact on individual class definitions. This property can be exploited to develop compressed TM models, enabling energy-efficient and high-throughput inferences for machine learning (ML) applications.

We introduce a training approach that incorporates excluded automata states to sparsify TM logic patterns in both positive and negative clauses. This exclusion is iterative, ensuring that highly class-correlated (and therefore significant) literals are retained in the compressed inference model, ETHEREAL, to maintain strong classification accuracy. Compared to standard TMs, ETHEREAL TM models can reduce model size by up to 87.54\%, with only a minor accuracy compromise. We validate the impact of this compression on eight real-world Tiny machine learning (TinyML) datasets against standard TM, equivalent Random Forest (RF) and Binarized Neural Network (BNN) on the STM32F746G-DISCO platform. Our results show that ETHEREAL TM models achieve over an order of magnitude reduction in inference time (resulting in higher throughput) and energy consumption compared to BNNs, while maintaining a significantly smaller memory footprint compared to RFs.
\end{abstract}

\keywords{Tsetlin Machine, Machine Learning, Model Compression, TinyML}

\maketitle

\section{Introduction} \label{sec:intro}
The ever increasing demand for deploying machine learning (ML) in low-energy, resource-constrained edge applications presents a significant challenge for deep neural network (DNN) implementations due to their high computational demands. This has led to efforts to identify alternative low-complexity ML algorithms. One such alternative is the Tsetlin Machine (TM), which is a novel ML algorithm that has been demonstrated with lower complexity than DNN, while achieving comparable accuracy across a range of ML datasets and exhibiting inherent interpretability \cite{granmo2018tsetlin}. A TM marks a fundamental shift from DNN by relying primarily on logic operations, which for example could outperform a multi-layer neural network (NN) in terms of accuracy \cite{tang2024adatm}, while eliminating hundreds of thousands of multiply-accumulate operations.

Figure \ref{fig:tm_structure} demonstrates a typical TM structure for supervised ML. The structure comprises three incremental processes:
\begin{itemize}
	\item [A.]\textit{Booleanization}: Before TM training and inference regimes, the input dataset is first expressed in the form of a set of literals, represented as Boolean data. These literals are derived through a data encoding process, known as Booleanization. A typical Booleanization process uses fixed or dynamic thresholds to generate Boolean literals as opposed to Binarized features from the raw data \cite{9923830}.
	\item [B.]\textit{Training}: Booleanized literals are given to a group of clauses, each learning a sub-pattern of some literals and performing AND operations to independently make a decision. Each clause learns these patterns through Tsetlin Automata (TAs), which decide whether a literal is included (above middle state) or excluded (below middle state), after a reinforcement learning process, see Section \ref{sec:learn} for further details. Half of all clauses have positive/negative polarity, capturing sub-patterns to support/oppose a classification. 
	\item [C.]\textit{Inference}: A binary classification is performed by a majority vote between the sum of outputs from positive and negative clauses. A multi-class classification requires as many pairs of positive-negative clauses as classes, where the overall classification is based on the one with the greatest class sum.
\end{itemize}

\input{TM_overview}

After training, TA array exhibits high sparsity; for example in the case of an MNIST dataset there are more than 99\% excludes. This property was leveraged to derive a compact model representation, REDRESS \cite{maheshwari2023redress}. The model only stores the information of includes as relative clauses and literal addressing. However, 
REDRESS, applied as a post-training compression, still follows the standard training process of the vanilla TM, resulting in a sparse form of training where the number of includes is not minimized, retaining less relevant or irrelevant context.
Reducing the number of includes is important in TMs, as their inherent sparse nature often incorporates literals with weak correlation to the target classes.  

It is possible to develop a more efficient TM by eliminating literals with weak correlations to a class, leading to minimal accuracy loss. 
{Though extensive research has been conducted on pruning weakly correlated features in DNNs, we emphasize that TM employs fundamentally different learning mechanism and data representation, and thus the pruning methods for DNNs are not applicable.}
In this work, we leverage the inherent interpretability of TM to identify weakly correlated literals, which are often included in both positive and negative clauses, due to their lack of strong association with a class. We propose a training approach to remove these literals, compressing TM models at the algorithm level, beyond REDRESS. This method, called \textbf{ETHEREAL}, enables \textbf{E}nergy-efficien\textbf{T}, \textbf{H}igh-throughput and accurate inf\textbf{E}rence through the practical implementation of a comp\textbf{RE}ssed tset\textbf{L}in m\textbf{A}chine. 

ETHEREAL introduces an additional exclusion process during training, to exclude literals shared by positive and negative clauses.
The exclusion is iteratively followed by standard training to restore important features. Results from eight real-world Tiny Machine Learning (TinyML) datasets show that ETHEREAL can realize up to an 87.54\% reduction in model size with a maximum accuracy loss of only 3.38\%, compared to a vanilla TM \cite{granmo2018tsetlin}. In some cases, accuracy even improves by eliminating some features that contribute noise. We use STM32F746G-DISCO micro-controller as the platform to implement ETHEREAL alongside REDRESS TM \cite{maheshwari2023redress}, a Random Forest (RF) and a Binarized Neural Network (BNN) \cite{courbariaux2016binarized, geiger2020larq}. The TM implementations can provide up to an order of magnitude reduction in inference time and energy compared to BNN, and 7$\times$ lower memory footprints than RF, while giving comparable accuracy. ETHEREAL further improves these design metrics, commensurate with the model size reductions achieved over REDRESS TMs. 

In this paper, we make the following key \textbf{\textit{contributions}}:
\begin{itemize}
	\item Empirical evidence revealing the inefficiency of vanilla TM in including less correlated (and thereby insignificant) literals.
	\item A training approach with additional exclusion, effectively compressing TM model and ensuring high accuracy.
	\item Validation with TinyML benchmarks on STM32 micro-controller, validating improved throughput, energy and memory usage produced by ETHEREAL. 
\end{itemize}


\section{TM Learning Dynamics} \label{sec:learn}
A TM is trained to capture the sub-pattern supporting or opposing a proposition by adjusting the TA states, which determine inclusion or exclusion of literals, driven by Type I and Type II feedback.

\begin{figure}[!htb]
\centering

\begin{tikzpicture}[font=\small]
\node [draw, thick, shape=rectangle, minimum width=0.8cm] at (0,-0.3) {$y$};
\draw []  (0,-0.5) to [bend left] (-0.2,-0.7);
\draw [] (-0.2,-0.7) -- (-1.8,-0.7);
\node[] at (-1,-0.5) {1};
\draw []  (-1.8,-0.7) to [bend right] (-2,-0.8);
\draw [] (-2,-0.8) -- (-2,-1.25);
\draw []  (0,-0.5) to [bend right] (0.2,-0.7);
\draw [] (0.2,-0.7) -- (1.8,-0.7);
\node[] at (1,-0.5) {0};
\draw []  (1.8,-0.7) to [bend left] (2,-0.8);
\draw [] (2,-0.8) -- (2,-1.25);

\draw []  (-2,-1.45) to [bend left] (-2.2,-1.65);
\draw [] (-2.2,-1.65) -- (-3.5,-1.65);
\node[] at (-2.85,-1.85) {positive};
\draw[] (-3.5,-1.65) to [bend right] (-3.7,-1.75);
\draw[] (-3.7,-1.75) -- (-3.7,-3);
\draw []  (-2,-1.45) to [bend right] (-1.8,-1.65);
\draw [] (-1.8,-1.65) -- (-0.5,-1.65);
\node[] at (-1.15,-1.85) {negative};
\draw[] (-0.5,-1.65) to [bend left] (-0.3,-1.75);
\draw[] (-0.3,-1.75) -- (-0.3,-2.6);
\node [draw, thick, shape=rectangle, minimum width=0.8cm,fill=white] at (-2,-1.2) {Clause polarity};

\draw [-stealth,dashed]  (-3.1,-2.8) to [bend right] (-1.35,-3.2);

\node [draw, thick, shape=rectangle, minimum width=0.8cm, rounded corners, fill=yellow!30!white] at (-3.1,-2.8) {\begin{tabular}[c]{@{}c@{}} \textbf{Type I} \\ feedback \end{tabular}};

\draw []  (2,-1.45) to [bend left] (1.8,-1.65);
\draw [] (1.8,-1.65) -- (0.5,-1.65);
\node[] at (1.15,-1.85) {positive};
\draw[] (0.5,-1.65) to [bend right] (0.3,-1.75);
\draw[] (0.3,-1.75) -- (0.3,-2.6);
\draw []  (2,-1.45) to [bend right] (2.2,-1.65);
\draw [] (2.2,-1.65) -- (3.5,-1.65);
\node[] at (2.95,-1.85) {negative};
\draw[] (3.5,-1.65) to [bend left] (3.7,-1.75);
\draw[] (3.7,-1.75) -- (3.7,-2.6);
\node [draw, thick, shape=rectangle, minimum width=0.8cm,fill=white] at (2,-1.2) {Clause polarity};

\draw [-stealth, dashed]  (3.1,-2.8) to [bend left] (1.35,-3.2);

\node [draw, thick, shape=rectangle, minimum width=0.8cm, rounded corners, fill=yellow!30!white] at (3.1,-2.8) {\begin{tabular}[c]{@{}c@{}} \textbf{Type I}  \\ feedback \end{tabular}};

\draw [-stealth, densely dashed] (0,-2.6) -- (0,-2.95);

\node [draw, thick, shape=rectangle, minimum width=0.8cm, rounded corners, fill=blue!20!white] at (0,-2.4) {\textbf{Type II} feedback};

\node [draw, densely dashed, thick, shape=rectangle, minimum width=0.8cm, rounded corners] at (0,-3.2) {with probability $\mathcal{P}$};

\end{tikzpicture}

\caption{TM feedback procedure, independently performed for each clause. For binary classification, $y$=1 or 0 suggests the sample belongs to the class or not, respectively; for multiclass classification and a TM for a certain class, $y$=1 or 0 suggests the sample belongs to the class or other classes, respectively.}
\label{fig:tm_feedbacks}

\Description[overall TM feedback]

\end{figure}
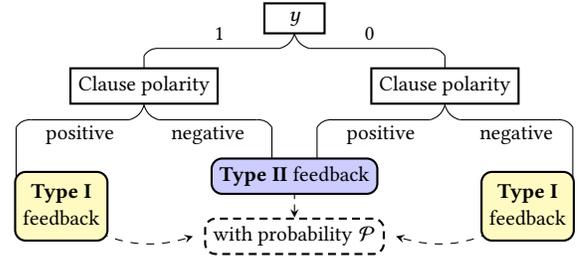

Figure \ref{fig:tm_feedbacks} explains the conditions under which each type of feedback is initiated. For a TA with 2N states, all TA states are initially set to either N or N+1 at random ($i.e.$, near the confusion state). During training, feedback is probabilistically activated for each datapoint; each specific type of feedback as well as the TAs being reinforced are determined by the training outcomes at the class and clause levels. 
Type I/II feedback is activated for all positive/negative clauses when $y$=1, while an opposite reaction occurs when $y$=0.

Figure \ref{fig:tm_feedbacks} shows both types of feedback are triggered with probability $\mathcal{P}$, determined by a hyperparameter $T$, as in (\ref{eq:feedback_prob}):
\vspace{-0.2cm}
\begin{IEEEeqnarray}{rCl}
	\mathcal{P} =\dfrac{T+(-1)^y\times\textrm{clip}(\sum\limits_{j=1}^{M} p_j C_j, -T, T)}{2T}
	\label{eq:feedback_prob}
\end{IEEEeqnarray}
where $M$ is the number of clauses; $p_j$ and $C_j$ are the polarity and output, respectively, for a specific clause. According to (\ref{eq:feedback_prob}), the farther the class sum is from $T$/$-T$ when $y$=1/0, the more likely the feedback is triggered, potentially calibrating more clauses to cast correct votes. On the other hand, feedback is withheld if the class sum becomes greater/smaller than $T$/$-T$, when $y$=1/0. Therefore, $T$ reveals the confidence in distinguishing between different classes. 

Figure \ref{fig:type_i_ii} illustrates the mechanism of both types of feedback.
In Type I feedback (Figure \ref{fig:type_i_ii} (a)), a clause that correctly supports or opposes the class (by producing an output of `1') is likely to include more literals that equal `1' at the datapoint.
This enables it to continue making the right decision by using a more fine-grained sub-pattern. On the other hand, the TA state of a literal equal to `0' is decreased to prevent it from overturning the correct output. 

Finally, a clause that fails to support the correct class (by producing an output of `0') may cause a false negative.
As a result, all TA of the clause are penalized by decreasing their states. 
In other words, Type I feedback combats false negatives by denying established sub-patterns and regenerating them in later learning process.

\definecolor{new_red}{rgb}{0.95,0.47,0.47}
\definecolor{new_green}{rgb}{0.85,0.98,0.77}

\begin{figure}[!htb]
\vspace{-0.5cm}
\centering
\subfloat[]{
\begin{tikzpicture}[font=\small]
\node [draw=none, shape=rectangle, minimum width=0.8cm, rounded corners, fill=yellow!30!white] at (0,0.6) {\textbf{Type I} Feedback};

\draw []  (0,-0.25) to [bend left] (-0.2,-0.45);
\draw [] (-0.2,-0.45) -- (-1.1,-0.45);
\node[] at (-0.65,-0.6) {1};
\draw []  (-1.1,-0.45) to [bend right] (-1.3,-0.55);
\draw [->] (-1.3,-0.55) -- (-1.3,-0.95);

\draw []  (0,-0.25) to [bend right] (0.2,-0.45);
\draw [] (0.2,-0.45) -- (1.1,-0.45);
\node[] at (0.65,-0.6) {0};
\draw []  (1.1,-0.45) to [bend left] (1.3,-0.55);
\draw [->] (1.3,-0.55) -- (1.3,-2.15);
\node [draw, thick, shape=rectangle, minimum width=0.8cm, fill=white] at (0,0) {Clause output};

\node [draw, thick, shape=rectangle, minimum width=0.8cm] at (-1.3,-1.18) {Literal};
\draw[->] (-1.3,-1.4) -- (-1.3,-2.15);
\node[] at (-1.45,-1.775) {1};
\draw[] (-1.3,-1.4) to [bend right] (-1.1,-1.6);
\draw[] (-1.1,-1.6) -- (1.1,-1.6);
\draw[] (1.1,-1.6) to [bend left] (1.3,-1.7);
\node[] at (0,-1.775) {0};

\node [draw, thick, shape=rectangle, minimum width=0.8cm, rounded corners, fill=new_green] at (-1.35,-2.8) {\begin{tabular}[c]{@{}c@{}} TA state \\ \textbf{increments} with \\ probability $\frac{s-1}{s}$ \end{tabular}};

\node [draw, thick, shape=rectangle, minimum width=0.8cm, rounded corners, fill=new_red!50!white] at (1.35,-2.8) {\begin{tabular}[c]{@{}c@{}} TA state \\ \textbf{decrements} with \\ probability $\frac{1}{s}$ \end{tabular}};

\end{tikzpicture}
}
\hspace{0.2cm}
\subfloat[]{
\begin{tikzpicture}[font=\small]
\node [draw=none, shape=rectangle, minimum width=0.8cm, rounded corners, fill=blue!20!white] at (0,0.6) {\textbf{Type II} Feedback};

\draw [->]  (0,-1.4) -- (0,-2.15);
\node[] at (-0.15,-1.775) {0};
\node [draw, thick, shape=rectangle, minimum width=0.8cm, fill=white] at (0,0) {Clause output};
	
\draw [->]  (0,-0.25) -- (0,-0.95);
\node[] at (-0.15,-0.6) {1};
\node [draw, thick, shape=rectangle, minimum width=0.8cm,fill=white] at (0,-1.2) {Literal};

\node [draw, thick, shape=rectangle, minimum width=2cm, minimum height=1.32cm, rounded corners, fill=new_green] at (0,-2.8) {\begin{tabular}[c]{@{}c@{}} TA state \\ \textbf{increments} \end{tabular}};
	
\end{tikzpicture}
}
\vspace{-0.2cm}
\caption{(a) Type I and (b) Type II feedback, where TA state remains unchanged for any other cases.}
\label{fig:type_i_ii}

\Description[Type I/II feedback]

\end{figure}
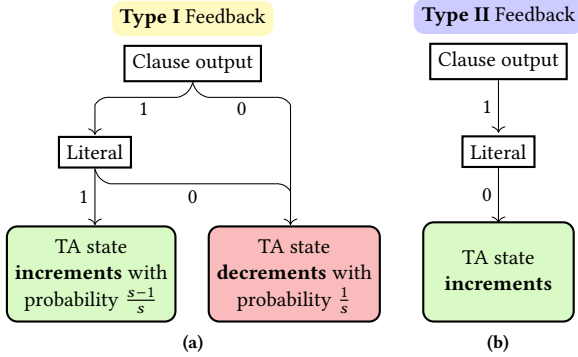

The probability of changing a TA state is determined by another hyperparameter, $s$, which indicates the probability of including a literal. The larger the value of $s$, the more/less likely a literal is to be included/excluded through Type I feedback. So far, the optimal values for both $T$ and $s$ are determined based on extensive trials aimed at achieving optimal accuracy~\cite{maheshwari2023redress, tarasyuk2023systematic}. 

If a clause incorrectly supports a class proposition, a false positive may occur. For instance, a positive/negative clause output is `1', when $y$=0/1. False positives are minimized by Type II feedback (Figure \ref{fig:type_i_ii} (b)). This type of feedback increases the TA states for the literals equaling `0', which potentially modifies the incorrect clause output of `1'. The TA states of a clause with output as `0' keeps unchanged, to avoid being trapped by local minima.

\section{ETHEREAL Model Compression} \label{sec:ethereal} 
\subsection{Literal Significance in Learning Dynamics}
The TM feedback mechanism given in Figure \ref{fig:type_i_ii}
ensures faster convergence during the training regime, through the interactions between both types of feedback. 
In addition, accuracy generally improves as more literals are included to capture fine-grained sub-patterns, as described in Section \ref{sec:learn}. However, this training process overlooks the significance or the correlation of individual literals to the target class. For example, a literal that consistently equals `1' does not provide useful information for classification, yet it can still be included in many clauses without adversely affecting accuracy.

We conduct an exploratory experiment to demonstrate how a TM model expands during training. In our experiment, the TM is trained to classify MNIST handwritten digits~\cite{deng2012mnist}, chosen as a case study for its simplicity in visualizing such an image classification task for our later analysis. We set the number of clauses per class, $T$ and $s$ to 100, 10 and 3, respectively, and Booleanize the dataset by applying a threshold of 75 to all grayscale values. Figure \ref{fig:TM_acc_size} shows resulting test accuracy and model size. As can be seen, the accuracy tends to increase with more training epochs, which is accompanied by a large increment on number of includes. This trend of model expansion is seen to hold across datasets and hyperparameters, as more TAs are included through random selection of automata reinforcements through $s$ and $T$ parameters explained above.

\pgfplotsset{width=3cm}

\pgfplotsset{select coords between index/.style 2 args={
		x filter/.code={
			\ifnum\coordindex<#1\def\pgfmathresult{}\fi
			\ifnum\coordindex>#2\def\pgfmathresult{}\fi
		}
}}

\begin{figure}[!htb]
\vspace{-0.6cm}
\centering
\subfloat[]{
		\begin{tikzpicture}[font=\normalsize, scale=0.95]
		\pgfplotsset{
			scale only axis,
		}
		
		\begin{axis}[
		height=3cm,
  	axis x line*=bottom,
	  axis y line*=left,
		ymax=96,
		ymin=91,
		xmin=1,
		xmax=50,
		xtick={10,20,30,40,50},
		grid=major,
		grid style={dashed,gray!50},
		ylabel = Accuracy (\%),
		xlabel = Training epoch number,
        ylabel style={yshift=-2.5ex}]	
		\addplot [very thick, blue] table [y=accuracy, x=epoch] {Data/vanillaTM_acc_size.dat};	
		\end{axis}
	\end{tikzpicture}
}
\subfloat[]{
		\begin{tikzpicture}[font=\normalsize, scale=0.95]
		\pgfplotsset{
			scale only axis,
		}

		\begin{axis}[
		height=3cm,
   	axis x line*=bottom,
	  axis y line*=left,
		ymax=40,
		ymin=10,
		xmin=1,
		xmax=50,
  	xtick={10,20,30,40,50},
  	grid=major,
		grid style={dashed,gray!50},
		ylabel = {\begin{tabular}[c]{@{}c@{}} Average number of \\ includes per clause \end{tabular}},
		xlabel = Training epoch number,
        ylabel style={yshift=-1.2ex}
		]
		\addplot [very thick, red] table [y=size, x=epoch] {Data/vanillaTM_acc_size.dat};
		\end{axis}
		
		\end{tikzpicture}
}
\vspace{-0.2cm}
	\caption{(a) Accuracy and (b) model size during training, for MNIST, as an example. Accuracy tends to increase with more epochs, accompanied by an increment in number of includes.} 
	\label{fig:TM_acc_size}

\Description[accuracy and size during training for MNIST]
    
\end{figure}
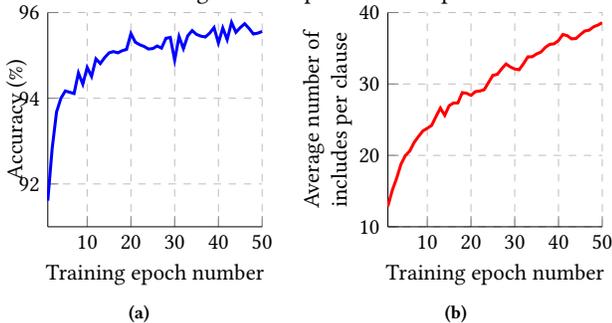


To investigate which literals are included during training, 
we visualize all complemented features in the image coordinate for a specific class (Figure \ref{fig:mnist_vanilla}). 
A notable observation from Figure \ref{fig:mnist_vanilla} (b) is that the features near digit outlines are more likely to be included in either positive or negative clauses, while those near the borders tend to be included in both types of clauses. This occurs because the border features do not effectively distinguish between classes, and can appear in samples from any class. Consequently, we conclude that insignificant literals are more likely to be included in both positive and negative clauses. Such observation is used to identify and exclude the insignificant literals, as described in Section \ref{sec:proposed}.     

\input{mnist_vanilla}

\vspace{-0.3cm}
\subsection{ETHEREAL Training} \label{sec:proposed}
The ETHEREAL training process consists of the following alternating steps,
repeated until the entire training is complete:
\begin{itemize}
	\item[1)] Conduct a specific number of standard training epochs, which is crucial for restoring any incorrectly excluded literals, as will be explained later.
	\item[2)] Identify all potentially insignificant literals, where a literal is considered as less insignificant if it is included in both positive and negative clauses.
	\item[3)] Exclude all potentially insignificant literals by adjusting their TA states, in exclusion process.
\end{itemize}

Specifically, a TM is initially trained for a certain number of epochs, using the standard training process, enabling it to identify preliminary sub-patterns. Subsequently, literals shared by positive and negative clauses (denoted by $l_i$) are identified, followed by an exclusion process, as shown in Figure \ref{fig:exclude_overview}. 

\definecolor{new_red}{rgb}{0.95,0.47,0.47}
\definecolor{new_green}{rgb}{0.85,0.98,0.77}

\definecolor{dark_red}{rgb}{0.78,0.28,0.34}
\definecolor{dark_green}{rgb}{0.09,0.30,0.28}

\definecolor{mypink}{rgb}{0.858, 0.188, 0.478}
\definecolor{babyblue}{rgb}{0.54,0.81,0.94}

\tikzset{
	node distance = 0pt,
	BB/.style args={#1/#2/#3}{
		draw, semithick, 
		font=\small\linespread{0.84}\selectfont, align=center, 
		minimum width=#1, minimum height=#2, text depth=0.25ex, 
		fill=#3, outer sep=0pt}
} 

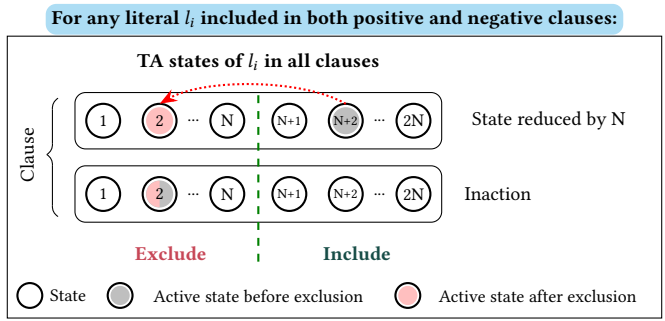
\begin{figure}[htp!]
\centering
\begin{tikzpicture}
[        GateCfg/.style={
	logic gate inputs={normal,normal},
	draw,
	thick,
	scale=1
	},
	cross/.style={path picture={ 
			\draw[black]
			(path picture bounding box.south east) -- (path picture bounding box.north west) (path picture bounding box.south west) -- (path picture bounding box.north east);
	}},
	scale=0.75, every node/.style={scale=0.75}
]





\begin{scope}[xshift=0cm,yshift=0cm]
\node[fill=babyblue!70!white,rounded corners] at (4.1,1.8) {\large{\textbf{For any literal $l_i$ included in both positive and negative clauses:}}};

\draw[draw=black,fill=white] (-1.7,-3.5) rectangle ++(11.6,5);

\node[] at (2.75,1.05) {\large{\textbf{TA states of $l_i$ in all clauses}}};

\draw[draw=black,fill=white,rounded corners] (-0.5,-0.5) rectangle ++(6.5,1);

\draw[thick,fill=white] (0,0) circle [radius=0.3] node (s1) {1};

\draw[thick,fill=white] (1,0) circle [radius=0.3] node (s2) {};
\draw[draw=none,fill=red!25!white] (1,0) circle [radius=0.24] node () {2};

\draw[thick,fill=white] (2.2,0) circle [radius=0.3] node (sn) {N};
\node[] at (1.6,0) {...};

\draw[thick,fill=white] (3.3,0) circle [radius=0.3] node (sn1) {\footnotesize{N+1}};

\draw[thick,fill=white] (4.3,0) circle [radius=0.3] node (sn2) {};
\draw[draw=none,fill=black!25!white] (4.3,0) circle [radius=0.24] node () {\footnotesize{N+2}};
\node[] at (4.9,0) {...};

\draw[thick,fill=white] (5.5,0) circle [radius=0.3] node (s2n) {2N};

\draw[thick, arrows={-Stealth[reversed,reversed]},thick,densely dotted,red] (4.3,0.3) to [out=150,in=30,looseness=0.7] (1,0.3);

\node[] at (7.9,0) {\large{State reduced by N}};

\end{scope}

\begin{scope}[xshift=0cm,yshift=-1.3cm]

\draw[draw=black,fill=white,rounded corners] (-0.5,-0.5) rectangle ++(6.5,1);

\draw[thick,fill=white] (0,0) circle [radius=0.3] node (s1) {1};

\draw[thick,fill=white] (1,0) circle [radius=0.3] node (s2) {};
\draw[draw=none,fill=black!25!white] (1,0) circle [radius=0.24] node () {};

\draw[fill=red!25!white,draw=none] (1,0.235) arc (90:270:0.235) -- cycle;
\node[] at (1,0) {2};

\draw[thick,fill=white] (2.2,0) circle [radius=0.3] node (sn) {N};
\node[] at (1.6,0) {...};

\draw[thick,fill=white] (3.3,0) circle [radius=0.3] node (sn1) {\footnotesize{N+1}};

\draw[thick,fill=white] (4.3,0) circle [radius=0.3] node (sn2) {\footnotesize{N+2}};
\node[] at (4.9,0) {...};

\draw[thick,fill=white] (5.5,0) circle [radius=0.3] node (s2n) {2N};

\node[] at (7,0) {\large{Inaction}};

\node[] at (1.2,-1.05) {\textcolor{dark_red}{\large{\textbf{Exclude}}}};
\node[] at (4.5,-1.05) {\textcolor{dark_green}{\large{\textbf{Include}}}};
\draw[thick,dashed,green!50!black] (2.75,-1.2) -- (2.75,1.8);

\begin{scope}[xshift=-0.5cm,yshift=0cm]
\draw[thick,fill=white] (-0.8,-1.8) circle [radius=0.22] node () {};
\node[] at (-0.12,-1.8) {State};
\draw[thick,fill=white] (0.8,-1.8) circle [radius=0.22] node () {};
\draw[draw=none,fill=black!25!white] (0.8,-1.8) circle [radius=0.17] node () {};
\node[] at (3.25,-1.8) {Active state before exclusion};
\draw[thick,fill=white] (5.9,-1.8) circle [radius=0.22] node () {};
\draw[draw=none,fill=red!25!white] (5.9,-1.8) circle [radius=0.17] node () {};
\node[] at (8.2,-1.8) {Active state after exclusion};
\end{scope}

\end{scope}

\begin{scope}[xshift=0.2cm,yshift=0cm]
\node[rotate=90] at (-1.55,-0.6) {\large{Clause}};
\draw[] (-1.2,-0.6) to [bend left] (-1.1,-0.8);
\draw[] (-1.2,-0.6) to [bend right] (-1.1,-0.4);
\draw[] (-1.1,-0.4) -- (-1.1,0.2);
\draw[] (-1.1,0.2) to [bend left] (-1,0.4);
\draw[] (-1.1,-0.8) -- (-1.1,-1.5);
\draw[] (-1.1,-1.5) to [bend right] (-1,-1.7);
\end{scope}

\end{tikzpicture}

\caption{ETHEREAL exclusion process.}
\label{fig:exclude_overview}

\Description[overall ETHEREAL process]

\end{figure}

For any clause including $l_i$, the TA states of $l_i$ are reduced by N, assuming each TA has a total of 2N states. This scheme ensures that $l_i$ is completely excluded from all clauses, while preserving its relative TA state: 
a ``strong include" (indicating a relatively high TA state) becomes a ``weak exclude", and a ``weak include" becomes a ``strong exclude".
For clauses that do not contain $l_i$, TA states remain unchanged, allowing the excluded $l_i$ with a TA state near the middle to possibly be restored in later training epochs.
Finally, literals that appear only in positive or negative clauses remain as they are, treated as crucial literals with strong correlation to target. 

A relatively significant literal may be predominantly included in one of the two types of clauses, but also appears in the other. Such literals may be improperly excluded. However, they are expected to have many candidate clauses with TA states near the middle state, allowing them to be restored after one or more training epochs. 


Figure \ref{fig:exclude_training} depicts the compressed TM model for MNIST. As can be seen, the model improves accuracy with fluctuations, while ETHEREAL results in a slower growth in the number of includes compared to the vanilla TM.
This gives a 46.6\% reduction in model size, with only a slight accuracy drop. An even greater reduction in model size is expected with additional training epochs.


\pgfplotsset{width=7cm,compat=1.5}

\pgfplotsset{select coords between index/.style 2 args={
		x filter/.code={
			\ifnum\coordindex<#1\def\pgfmathresult{}\fi
			\ifnum\coordindex>#2\def\pgfmathresult{}\fi
		}
}}

\begin{figure}[htp!]
\centering
\subfloat[]{
		\begin{tikzpicture}[font=\small, scale=1, every node/.style={scale=1}]
		\pgfplotsset{
			scale only axis,
		}
		
		\begin{axis}[
		height=4cm,
		axis x line*=bottom, 
		axis y line*=left,
		ymax=96,
		ymin=91,
		xmin=1,
		xmax=50,
		xtick={1,10,20,30,40,50},
		grid=major,
		grid style={dashed,gray!50},
		ylabel = Accuracy (\%),
		xlabel = Training epoch number,
		clip=false,
		legend style={at={(axis cs:7,92)},anchor=west,font=\scriptsize},
		ylabel style={yshift=-0.2cm},
		]	
		\addplot [semithick,blue] table [y=accuracy, x=epoch] {Data/vanillaTM_acc_size.dat};
		\addlegendentry{Vanilla TM}
		\addlegendimage{ultra thin, red, mark=triangle*}\addlegendentry{ETHEREAL after training}
		\addlegendimage{ultra thin, red, mark=*}\addlegendentry{ETHEREAL after exclusion}
		\addplot [ultra thin, red] table [y=accuracy, x=epoch]{Data/excludeTM_acc_size.dat};
		\addplot [only marks, red, mark=triangle*, mark size=1.5pt, ultra thin] table [y=accuracy, x=epoch]{Data/excludeTM_acc_size_OnlyRetrain.dat};	
		\addplot [only marks, red, mark=*, mark size=1.5pt, ultra thin] table [y=accuracy, x expr=\thisrow{epoch}+1]{Data/excludeTM_acc_size_OnlyExclude.dat};
		\draw[fill=white,thick, densely dotted] (axis cs:35.3,91.2) rectangle (axis cs:51,93.7);
		\draw[thick, densely dotted] (axis cs:35.5, 94.3) rectangle (axis cs:41.5,95.1);
		\coordinate[] (pt) at (axis cs:36.5,94.4);		
		\end{axis}
		
		\node[pin=280:{%
		\begin{tikzpicture}[baseline,trim axis left,trim axis right,scale=0.25]
		\begin{axis}[
		xmin=36,xmax=41,
		xtick={36,37,38,39,40,41},
		ymin=94.4,ymax=95,
		ytick={94.4,94.7,95},
		enlargelimits,
		grid=major,
		grid style={dashed,gray!50},
		ticklabel style = {font=\Huge}
		]
		\addplot [very thick,red] table [y=accuracy, x=epoch]{Data/excludeTM_acc_size.dat};
		\addplot [only marks, red, mark=triangle*, mark size=6pt] table [y=accuracy, x=epoch]{Data/excludeTM_acc_size_OnlyRetrain.dat};	
		\addplot [only marks, red, mark=*, mark size=6pt] table [y=accuracy, x expr=\thisrow{epoch}+1]{Data/excludeTM_acc_size_OnlyExclude.dat};
		\end{axis}
		\end{tikzpicture}%
		}] at (pt) {};
	
		\end{tikzpicture}
}

\subfloat[]{
	\begin{tikzpicture}[font=\small, scale=1, every node/.style={scale=1}]
	\pgfplotsset{
		scale only axis,
	}

	\begin{axis}[
	height=4cm,
	axis x line*=bottom,
	axis y line*=left,
	ymax=40,
	ymin=8,
	xmin=1,
	xmax=50,
	grid=major,
	grid style={dashed,gray!50},
	ytick={10,20,30,40},
	xtick={1,10,20,30,40,50},
	ylabel = {\begin{tabular}[c]{@{}c@{}} Average number of \\ includes per clause \end{tabular}},
	xlabel = Training epoch number,
	clip=false,
	legend style={at={(axis cs:2,38)},anchor=west,font=\scriptsize},
	ylabel style={yshift=-0.2cm}
	]		
	\addplot [semithick, blue] table [y=size, x=epoch] {Data/vanillaTM_acc_size.dat};
	\addlegendentry{Vanilla TM}
	\addlegendimage{ultra thin, red, mark=triangle*}\addlegendentry{ETHEREAL after training}
	\addlegendimage{ultra thin, red, mark=*}\addlegendentry{ETHEREAL after exclusion}
	\addplot [ultra thin, red] table [y=size, x=epoch]{Data/excludeTM_acc_size.dat};
	\addplot [only marks, red, mark=triangle*, mark size=1.5pt, ultra thin] table [y=size, x=epoch]{Data/excludeTM_acc_size_OnlyRetrain.dat};	
	\addplot [only marks, red, mark=*, mark size=1.5pt, ultra thin] table [y=size, x expr=\thisrow{epoch}+1]{Data/excludeTM_acc_size_OnlyExclude.dat};
	\draw[thick, densely dotted] (axis cs:35.5, 15.5) rectangle (axis cs:41.5,21.5);
	\draw[fill=white,thick, densely dotted] (axis cs:39,22.5) rectangle (axis cs:53.5,35);
	\coordinate[] (pt) at (axis cs:36,20);
	\end{axis}

	\node[pin=20:{%
	\begin{tikzpicture}[baseline,trim axis left,trim axis right,yscale=0.2, xscale=0.25]
	\begin{axis}[
	xmin=36,xmax=41,
	xtick={36,37,38,39,40,41},
	ymin=16.5,ymax=20.5,
	ytick={18,20},
	enlargelimits,
	grid=major,
	grid style={dashed,gray!50},
	ticklabel style = {font=\Huge}
	]
	\addplot [very thick,red] table [y=size, x=epoch]{Data/excludeTM_acc_size.dat};
	\addplot [only marks, red, mark=triangle*, mark size=6pt] table [y=size, x=epoch]{Data/excludeTM_acc_size_OnlyRetrain.dat};	
	\addplot [only marks, red, mark=*, mark size=6pt] table [y=size, x expr = \thisrow{epoch}+1]{Data/excludeTM_acc_size_OnlyExclude.dat};
	\end{axis}
	\end{tikzpicture}%
	}] at (pt) {};

	\end{tikzpicture}

}
	\caption{(a) Accuracy and (b) model size, for vanilla and ETHEREAL TMs, for MNIST. ETHEREAL causes an accuracy drop of 0.8\% (from 95.8\% to 95\%) but realizes a 46.6\% reduction in model size, where the average number of includes per clause is reduced from 36.3 to 19.4.}
	\label{fig:exclude_training}

\Description[accuracy and size during training for ETHEREAL]
    
\end{figure}
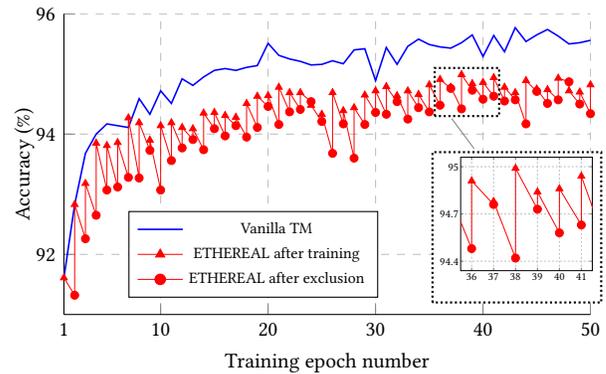
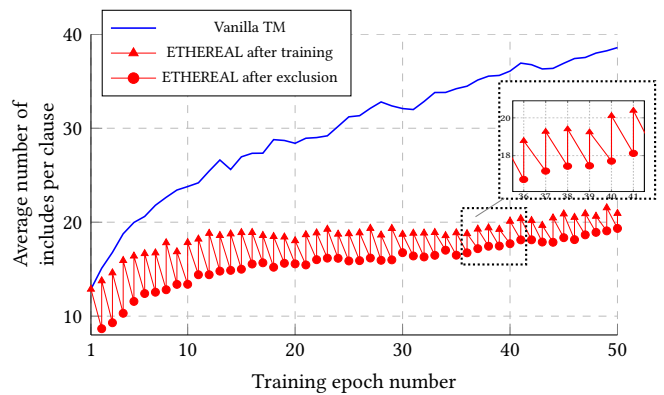

\begin{figure}[htp!]
	\setlength{\fboxsep}{3mm} 
	\setlength{\tabcolsep}{0pt}
	\centering

	\begin{tikzpicture}[scale=0.17,every node/.style={scale=0.17}]
	\node[] at (0,0) {
		\begin{tabular}{*{28}{G}}
0&0&0&0&0&0&0&0&0&0&0&0&0&1&0&0&0&0&1&0&0&0&0&0&0&0&0&0\\
0&0&0&0&1&0&0&0&0&0&0&0&0&0&0&0&0&0&0&0&0&0&0&0&0&0&0&0\\
0&0&0&0&0&0&0&0&0&1&0&0&0&0&0&0&0&0&0&1&0&1&0&0&0&0&0&0\\
1&0&0&0&1&0&1&0&0&0&0&0&0&0&0&0&0&1&0&1&1&0&2&0&0&0&1&0\\
0&0&0&0&0&0&0&0&0&0&0&0&0&0&0&0&0&0&1&2&0&0&1&0&0&0&0&1\\
0&0&0&0&0&0&1&0&0&0&0&0&0&0&0&0&0&1&0&1&1&1&0&1&0&0&0&0\\
1&0&0&0&0&0&1&2&0&0&0&0&1&0&0&0&0&0&0&0&0&1&0&1&0&0&0&0\\
0&0&0&1&0&0&0&0&0&0&0&0&0&0&0&0&0&0&2&0&0&1&0&1&0&0&0&1\\
0&0&0&0&1&1&0&0&0&0&0&0&0&0&0&1&0&0&0&0&0&3&1&0&0&0&0&0\\
0&0&0&0&0&1&0&3&0&3&0&0&0&0&0&0&0&0&0&0&0&2&0&0&0&0&0&0\\
0&0&0&0&0&0&1&0&0&1&2&3&0&0&1&1&0&0&1&0&1&0&1&0&0&0&0&0\\
0&0&0&0&0&0&2&0&1&1&4&3&6&3&2&2&0&0&0&0&0&2&0&0&0&0&0&0\\
0&0&0&1&0&1&0&2&0&4&2&4&5&4&3&0&1&0&0&0&0&3&0&1&0&0&1&0\\
0&0&0&1&0&3&1&0&0&3&3&0&0&1&0&0&0&0&0&1&1&1&0&0&1&0&0&0\\
0&0&0&0&1&0&1&2&0&0&2&2&0&0&0&0&0&0&0&1&1&0&0&0&0&0&0&0\\
0&0&0&0&0&1&2&1&1&1&0&0&0&0&0&0&0&0&1&1&2&2&0&1&0&0&0&0\\
0&0&0&0&1&1&0&1&1&0&0&0&0&0&0&0&0&1&0&0&0&0&0&0&0&0&0&0\\
0&0&0&0&0&1&0&0&0&0&0&0&0&0&0&0&0&0&0&1&0&0&0&0&1&0&1&0\\
0&0&0&0&0&0&0&0&0&0&0&0&0&0&0&0&0&0&0&0&0&0&0&0&0&0&0&0\\
0&0&0&1&0&0&1&0&0&0&0&0&0&0&0&0&0&0&0&0&0&0&0&0&0&0&1&0\\
0&1&1&0&0&0&0&0&0&0&0&0&0&0&0&0&0&0&0&0&0&0&0&0&0&0&0&0\\
0&0&0&0&0&0&0&0&0&0&0&0&0&1&1&1&0&2&0&0&0&0&1&0&0&0&0&1\\
0&0&0&0&1&0&0&0&0&0&0&0&0&0&1&1&0&0&0&0&0&1&0&0&0&0&0&0\\
0&0&0&0&0&0&1&5&0&2&0&2&0&2&5&3&0&0&1&0&0&0&0&0&0&0&0&0\\
0&0&0&0&0&1&0&3&3&4&5&4&5&5&0&1&0&0&0&0&0&0&0&0&0&0&0&0\\
0&1&0&0&0&0&2&0&5&0&3&5&1&0&0&1&0&0&0&1&0&0&0&0&0&0&0&0\\
0&0&1&0&0&0&0&0&0&0&0&0&0&0&0&0&0&0&0&0&0&0&0&1&0&0&1&0\\
0&0&0&0&0&0&0&1&0&0&0&0&0&0&0&0&0&0&0&0&0&0&0&1&0&0&0&0\\
		\end{tabular}
	};
\draw[draw=gray, dashed] (-8.5,-8.5) rectangle ++(17,17);
\draw[-{Straight Barb[angle=60:2pt 3]}] (-9, -8.5) -- (9, -8.5);
\node[] at (0, -9.5) {\HUGE{Horizontal position}};
\draw[-{Straight Barb[angle=60:2pt 3]}] (-8.5, -9) -- (-8.5, 9);
\node[rotate=90] at (-9.5, 0) {\HUGE{Vertical position}};

\node[] at (20,0) {
	\begin{tabular}{*{28}{G}}
2&13&3&7&0&0&9&7&8&6&0&2&2&4&7&0&3&7&3&4&0&1&6&1&3&4&3&3\\
3&5&4&2&3&1&5&2&1&0&5&2&2&5&4&3&0&3&0&2&1&4&2&4&1&6&0&2\\
3&3&8&5&7&12&0&10&3&0&5&1&3&2&0&2&3&0&2&2&2&1&1&5&5&3&6&4\\
4&4&2&8&2&3&3&4&2&4&4&3&2&2&1&0&1&0&0&0&1&0&0&1&4&2&3&2\\
0&3&12&13&1&0&2&2&3&5&3&0&3&1&3&0&1&1&0&0&0&0&0&1&1&0&4&9\\
5&1&1&6&1&0&2&2&4&0&3&2&0&0&1&0&2&0&0&0&0&0&2&1&0&5&9&6\\
5&2&2&2&4&2&2&0&2&1&1&0&0&0&1&0&0&0&0&0&0&0&0&0&1&3&8&1\\
5&0&2&2&1&1&2&0&2&0&1&0&0&0&0&0&0&0&0&0&0&0&0&0&0&0&0&4\\
3&1&0&0&0&0&1&0&1&1&0&0&0&0&0&0&0&0&0&0&0&0&0&2&0&2&1&9\\
2&7&5&1&1&2&0&0&0&0&0&0&0&0&0&0&0&0&0&0&0&0&0&0&1&1&5&0\\
2&4&1&0&2&1&1&0&0&0&0&0&0&0&0&0&0&0&0&0&0&0&0&0&0&4&4&5\\
5&4&5&0&3&1&0&0&0&0&0&0&0&0&0&0&0&0&0&0&0&0&1&1&0&2&2&3\\
5&3&1&0&0&0&3&0&1&0&0&0&0&0&0&0&0&0&0&0&0&0&1&0&2&5&0&11\\
6&1&2&2&0&0&1&0&1&0&0&1&0&0&0&0&0&0&0&0&0&0&1&3&0&0&4&3\\
3&5&5&2&0&1&1&1&0&0&0&0&1&0&0&0&0&0&0&0&0&0&0&2&0&5&6&2\\
1&2&6&1&2&2&1&0&0&0&1&0&0&0&0&0&0&0&0&0&0&0&0&0&4&6&12&0\\
4&0&1&1&0&0&2&0&2&0&2&1&2&0&0&1&0&0&0&0&0&0&2&3&7&2&16&7\\
9&2&4&1&0&5&0&1&0&0&1&1&1&1&0&2&0&0&0&0&2&4&2&5&10&20&2&6\\
12&4&1&0&2&2&0&2&4&2&0&1&1&3&1&0&0&0&0&0&0&2&4&6&3&4&8&3\\
2&6&1&4&1&0&2&2&2&2&1&1&0&2&1&1&0&0&0&2&1&9&4&12&9&6&12&3\\
0&8&0&1&5&0&3&1&1&1&0&0&0&0&0&0&0&0&2&5&5&15&10&14&2&7&3&0\\
2&2&0&0&2&1&0&0&1&0&0&0&0&0&0&0&0&0&1&4&6&8&12&8&9&3&1&2\\
0&10&2&0&0&4&1&2&3&1&0&0&0&0&0&0&2&1&4&8&6&7&5&5&5&1&4&5\\
2&1&8&8&1&0&0&0&0&0&0&0&0&0&0&0&1&0&3&2&0&3&2&4&2&1&1&5\\
12&4&1&1&1&0&2&0&0&0&0&0&0&0&0&0&2&0&2&1&2&1&1&4&1&6&4&3\\
0&7&1&12&8&1&0&1&0&1&0&0&0&0&1&1&0&0&0&1&2&3&0&3&0&0&3&8\\
7&3&2&1&9&1&0&1&2&5&2&1&0&3&0&0&0&3&2&1&1&6&4&1&0&3&6&0\\
2&3&6&1&9&0&5&1&1&8&0&1&1&6&2&0&12&0&2&1&6&0&2&8&10&10&4&1\\
	\end{tabular}
};
\draw[draw=gray, dashed] (11.6,-8.5) rectangle ++(17,17);
\draw[-{Straight Barb[angle=60:2pt 3]}] (11.1, -8.5) -- (29.1, -8.5);
\node[] at (20.1, -9.5) {\HUGE{Horizontal position}};
\draw[-{Straight Barb[angle=60:2pt 3]}] (11.6, -9) -- (11.6, 9);
\node[rotate=90] at (10.6, 0) {\HUGE{Vertical position}};

\node[] at (0,9.5) {\SUPERHUGE{\textbf{Positive clauses}}};
\node[] at (20.1,9.5) {\SUPERHUGE{\textbf{Negative clauses}}};
\node[] at (32,0) {
	\begin{tabular}{*{28}{G}}
	20&20\\
	19&19\\
	18&18\\
	17&17\\
	16&16\\
	15&15\\
	14&14\\
	13&13\\
	12&12\\
	11&11\\
	10&10\\
	9&9\\
	8&8\\
	7&7\\
	6&6\\
	5&5\\
	4&4\\
	3&3\\
	2&2\\
	1&1\\
	0&0\\
	\end{tabular}
};
\draw[draw=black] (31.4,-6) rectangle ++(1.1,12.3);
\draw[] (31.4,-6) -- (33.5, -6);
\node[] at (34.5,-6) {\HUGE{0}};
\draw[] (31.4,6.3) -- (33.5, 6.3);
\node[] at (34.5,6.3) {\HUGE{20}};
\draw[] (32.5,0.15) -- (33.5, 0.15);
\node[] at (34.5, 0.15) {\HUGE{10}};

\node[rotate=-90] at (37, 0.15) {\HUGE{Number of includes}};
	\end{tikzpicture}
			
	\caption{Number of includes for all complemented features, represented in a 28$\times$28 image coordinate for MNIST digit `2', after 50 epochs of ETHEREAL training. Comparing this figure with Fig. \ref{fig:mnist_vanilla} (b), the less significant features near the border are largely eliminated, especially from the positive clauses. Some significant features are still visible, such as those at the image center for positive clause case and in the bottom right corner for negative clause case.}
	\label{fig:mnist_diet}

\Description[interpretable results for ETHEREAL]
    
\end{figure}
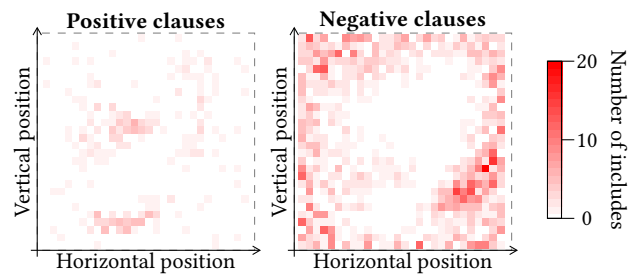

In Figure \ref{fig:mnist_diet}, we visualize the complemented features in the image coordinate for the ETHEREAL TM. The less significant features are largely excluded, while the more significant ones are retained, ensuring minimal loss in accuracy.

\section{Evaluation} \label{sec:eva}
\subsection{Experimental Setup}
\begin{figure}[!htb]
\vspace{-0.3cm}
\centering
\begin{tikzpicture}[font=\small]
\draw[-] (2,-0.7) -- (2,-1.1);
\node [draw, thick, shape=rectangle, minimum width=0.8cm, rounded corners, fill=white] at (2,-0.7) {\textbf{Raw dataset}};
\draw[] (2,-1.1) to [bend left] (1.9,-1.2);
\draw[] (1.9,-1.2) -- (1,-1.2);
\draw[] (1,-1.2) to [bend right] (0.9,-1.3);
\draw[-stealth] (0.9,-1.3) -- (0.9,-1.45); 

\draw[draw=none, fill=gray, opacity=0.5] (-1.5,-1.3) rectangle ++(3.2, -1.6);
\node [draw=none, fill=gray!50!white, shape=rectangle, minimum width=0.8cm] at (-0.43,-1.1) {\textbf{Booleanization}};

\draw[dashed, rounded corners, semithick] (2, -1.3) rectangle ++(4.9, -3.7);
\node [] at (5.7, -1.1) {Hyperparameters};

\draw[-stealth] (0,-1.7) -- (0,-2.25); 

\node [draw, thick, shape=rectangle, minimum width=0.8cm, rounded corners, fill=white] at (0,-1.7) {Quantile binning};
\draw[-stealth, dashed] (4.5, -1.7) -- (1.1, -1.7);
\node[trapezium, draw, thick, minimum width=0.3cm, trapezium left angle=75, trapezium right angle=105, trapezium stretches=true, minimum height=0.3cm, fill=white] at (4.5,-1.7) {Number of bins};

\draw[-stealth] (0,-2.7) -- (0,-3.75); 

\node [draw, thick, shape=rectangle, minimum width=0.8cm, rounded corners, fill=white] at (0,-2.5) {Encoding};
\draw[-stealth, dashed] (4.5, -2.5) -- (0.65, -2.5);
\node[trapezium, draw, thick, minimum width=0.3cm, trapezium left angle=75, trapezium right angle=105, trapezium stretches=true, minimum height=0.3cm, fill=white] at (4.5,-2.5) {One-hot or thermometer};

\draw[-stealth] (0,-4) -- (0,-5.25); 
\node [draw, thick, shape=rectangle, minimum width=0.8cm, rounded corners, fill=white] at (0,-4) {\textbf{Off-platform training}};
\draw[-stealth, dashed] (3, -4) -- (1.5, -4);
\node[trapezium, draw, thick, minimum width=0.3cm, trapezium left angle=75, trapezium right angle=105, trapezium stretches=true, minimum height=0.3cm, fill=white] at (4.5,-4) {\begin{tabular}[c]{@{}c@{}} Total number of epochs, \\ number of clauses, $T$, $s$, \\ number of epochs after \\ each exclusion  \end{tabular}};

\draw[-stealth] (0,-5.7) -- (2.95,-5.7); 
\node [draw, thick, shape=rectangle, minimum width=0.8cm, rounded corners, fill=white] at (0,-5.7) {\begin{tabular}[c]{@{}c@{}} \textbf{Model exporting}: \\ REDRESS ecnoding \cite{maheshwari2023redress} \end{tabular}};

\node [draw, thick, shape=rectangle, minimum width=0.8cm, rounded corners] at (4.5,-5.7) {\textbf{On-platform inference}};

\end{tikzpicture}
\vspace{-0.6cm}
\caption{ML pipeline used for generating TM models for experimental validations.}
\label{fig:pipeline}

\Description[design pipeline]

\end{figure}
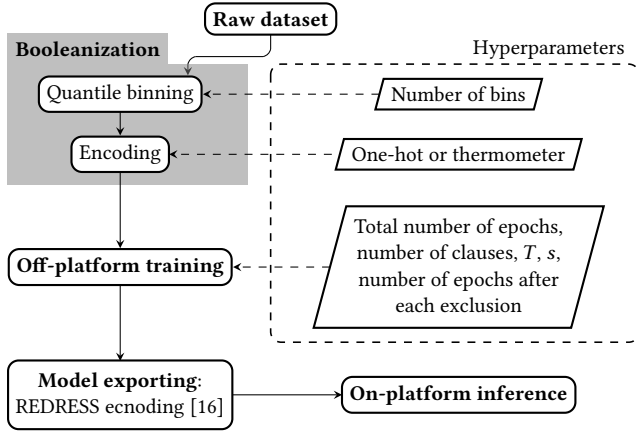

To validate the proposed inference model, a ML pipeline (Figure \ref{fig:pipeline}) is applied to produce both vanilla and ETHEREAL TM models, encoded with REDRESS \cite{maheshwari2023redress} and deployed on STM32F746G-DISCO micro-controller via Micropython. 

Eight real-world TinyML datasets (Table \ref{tab:dataset}) are selected from \cite{banbury2020benchmarking}, including electromyography (EMG) based gesture recognition \cite{lobov2018latent}, gas sensor array drift \cite{rodriguez2014calibration}, gesture phase segmentation (GPS) \cite{madeo2013gesture}, human activity recognition (HAR) \cite{anguita2013public}, mammographic mass \cite{elter2007prediction}, sensorless drive diagnosis \cite{dataset_for_sensorless_drive_diagnosis_325}, sport activity \cite{altun2010comparative}, and statlog (vehicle silhouette) \cite{mowforth1987}. 
To the best of our knowledge, there has been no prior work exploring TM models on these datasets. 
{Consequently, we determine all hyperparameters through trial and error, to achieve TM models with accuracy comparable to other reported ML algorithms \cite{karnam2022emghandnet, rodriguez2014calibration, madeo2013gesture, anguita2013public, elter2007prediction, jiang2016fault, altun2010comparative, king1995statlog}.
By definition, it is possible to further improve accuracy with more clauses, at the cost of greater computational resources \cite{granmo2018tsetlin,tarasyuk2023systematic}.} 
Relevant information about the datasets and hyperparameters is provided in Table \ref{tab:dataset}. Additional details and the source code for the entire pipeline are openly available at: \url{https://github.com/nsd5g13/TM4TinyML}. 

\begin{table}[!htb]
	\centering
	\caption{Dataset and hyperparameters configuration.}
	\label{tab:dataset}
\vspace{-0.2cm}	
\resizebox{\linewidth}{!}
{
\begin{threeparttable}
	
	\begin{tabular}{l|cccccc}
	\hline 
	 & \textbf{Classes} & \textbf{Features} & \textbf{Literals} & \textbf{Epochs} & (\textbf{Clauses}\tnote{*}, $T$, $s$) \\ \hline
	 EMG & 8 & 160 & 320 & 200 & (300, 14, 7.5) \\
	 Gas sensor & 6 & 128 & 256 & 200 & (300, 12, 10)\\
	 GPS & 5 & 18 & 360 & 250 & (500, 25, 9)\\
	 HAR & 6 & 560 & 1120 & 250 & (200, 14, 6)\\
	 Mammographic mass & 2 & 5 & 30 & 100 & (50, 7 ,3)\\
	 Sensorless drive & 11 & 48 & 288 & 100 & (300, 15, 10)\\
	 Sport activity & 19 & 45 & 90 & 50 & (150, 12 ,4)\\
	 Statlog & 4 & 18 & 720 & 100 & (300, 16, 3)\\
	 \hline

	\end{tabular}

	\begin{tablenotes}
	\scriptsize
	\item[*] Number of clauses per class
	
	\end{tablenotes}

\end{threeparttable}
}

\end{table}

\vspace{-0.5cm}
\subsection{Off-Platform Evaluation}
For off-platform evaluation, we assess model complexity and accuracy.
For the vanilla and ETHEREAL TMs, we report the best test accuracy achieved in the total number of epochs along with the model sizes corresponding to this accuracy, in Table \ref{tab:acc_size}. As can be seen, ETHEREAL significantly decreases the number of includes by 39.29-87.54\%, except for mammographic mass and statlog, while resulting in a small reduction (0.78-3.38\%) in accuracy.
It is most notable that ETHEREAL achieves equal or even slightly improved accuracy with fewer literals for mammographic mass and statlog. 

\definecolor{new_red}{rgb}{0.95,0.47,0.47}
\definecolor{new_green}{rgb}{0.85,0.98,0.77}

\begin{table*}[!htb]
	\centering
	\caption{Comparisons between RFs, BNNs, vanilla and ETHEREAL TMs.}
	\label{tab:acc_size}
\vspace{-0.3cm}
\resizebox{\textwidth}{!}{%
	\begin{tabular}{|c||cc||cc||cc||cccc|}
	\hline
 	\multirow{2}{*}{} & \multicolumn{2}{c||}{\textbf{RF}} & \textbf{BNN: FC256} & \textbf{BNN: FC512} & \multicolumn{2}{c||}{\textbf{Vanilla TM}} & \multicolumn{4}{c|}{\textbf{ETHEREAL TM}} \\ \cline{2-11}
 	& {\begin{tabular}[c]{@{}c@{}} (Trees, \\ Depth) \end{tabular}} & {\begin{tabular}[c]{@{}c@{}} Accuracy \\ (\%) \end{tabular}} & {\begin{tabular}[c]{@{}c@{}} Accuracy \\ (\%) \end{tabular}} & {\begin{tabular}[c]{@{}c@{}} Accuracy \\ (\%) \end{tabular}} & {\begin{tabular}[c]{@{}c@{}} Accuracy \\ (\%) \end{tabular}} & {\begin{tabular}[c]{@{}c@{}} Includes \\ per clause \end{tabular}} & {\begin{tabular}[c]{@{}c@{}} Accuracy \\ (\%) \end{tabular}} & {\begin{tabular}[c]{@{}c@{}} Accuracy \\ change (\%) \end{tabular}} & {\begin{tabular}[c]{@{}c@{}} Includes \\ per clause \end{tabular}} & {\begin{tabular}[c]{@{}c@{}} Includes \\ change (\%) \end{tabular}} \\ \hline 
 	EMG & (25,14) & 77.66 & 81.10 & 81.24 & 85.95 & 9.23 & 84.09 & \cellcolor{new_red!50} -1.86 & 5.18 & \cellcolor{new_green!70} -43.88 \\
 	Gas sensor & (30,12) & 91.29 & 81.19 & 80.07 & 87.19 & 12.27 & 84.89 & \cellcolor{new_red!50} -2.30 & 7.22 & \cellcolor{new_green!70} -41.16 \\
  	GPS & (25,16) & 66.68 & 81.02 & 83.19 & 82.53 & 29.31 & 79.15 & \cellcolor{new_red!50} -3.38 & 17.12 & \cellcolor{new_green!70} -41.59 \\
  	HAR & (25,14) & 84.93 & 80.52 & 80.76 & 88.33 & 66.22 & 87.55 & \cellcolor{new_red!50} -0.78 & 8.25 & \cellcolor{new_green!70} -87.54 \\
  	Mammographic mass & (30,4) & 83.94 & 82.38 & 81.35 & 83.94 & 3.07 & 83.94 & \cellcolor{new_green!70} +0.00 & 1.67 & \cellcolor{new_green!70} -45.60 \\
  	Sensorless drive & (30,16) & 88.89 & 56.84 & 72.86 & 86.45 & 15.30 & 85.12 & \cellcolor{new_red!50} -1.33 & 8.58 & \cellcolor{new_green!70} -43.94 \\
  	Sport activity & (25,20) & 87.50 & 79.11 & 86.37 & 92.61 & 5.81 & 89.64 & \cellcolor{new_red!50} -2.97 & 3.53 & \cellcolor{new_green!70} -39.29 \\
  	Statlog & (25,14) & 75.29 & 71.76 & 72.94 & 81.18 & 4.76 & 82.35 & \cellcolor{new_green!70} +1.17 & 4.70 & \cellcolor{new_green!70} -1.36 \\
 	\hline
	\end{tabular}
}
\end{table*}

{The above results suggest that the performance of ETHEREAL is determined by the given features and target:}
for datasets with a large amount of noisy features, accuracy could remain the same or improve by retaining significant features and excluding noise. Conversely, for datasets that rely on interactions among numerous similarly significant features, ETHEREAL results in a slight decrease in accuracy as some features are excluded.
{The results are unrelated to the dataset or model scales, by comparing Tables \ref{tab:dataset} and \ref{tab:acc_size}.}

For each dataset, 
{we train a RF model using the raw features with the number of trees (5–30) and maximum depth (2–20) selected via grid search for the hightest test accuracy.}
We also train two single hidden layer BNN models using the Boolean features with 256 and 512 fully connected neurons (FC256 and FC512) using Larq \cite{geiger2020larq}. 
All the algorithms are capable of achieving comparable accuracy. As the accuracy of each algorithm can be improved with further tuning, we do not directly compare the accuracy; instead, we will evaluate them based on accuracy and other design metrics in Section \ref{sec:stm32}.

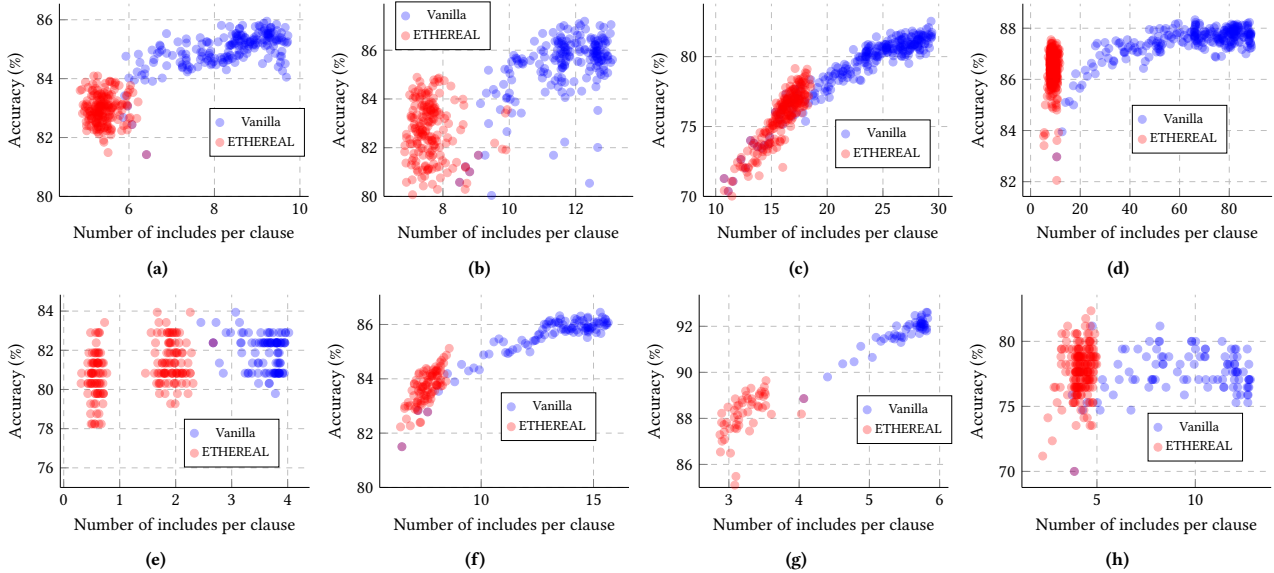
\begin{figure*}[!htb]
\vspace{-0.4cm}
\centering
\subfloat[]{
		\begin{tikzpicture}[font=\normalsize, scale=0.75]
		
		\begin{axis}[
		height=5cm,		
		width=6cm,
		axis x line*=bottom, 
		axis y line*=left,
		ymin=80,
		grid=major,
		grid style={dashed,gray!50},
		ylabel = Accuracy (\%),
		xlabel = Number of includes per clause,
		legend style={at={(axis cs:9,83)},anchor=north,font=\footnotesize},
		]	
		\addplot [only marks, blue, opacity=0.3] table [y=acc, x=size] {Data/vanilla_tradeoff_emg.dat};
		\addlegendentry{Vanilla}
		\addplot [only marks, red, opacity=0.3] table [y=acc, x=size] {Data/diet_tradeoff_emg.dat};
		\addlegendentry{ETHEREAL}	
		\end{axis}

		\end{tikzpicture}
}
\subfloat[]{
	\begin{tikzpicture}[font=\normalsize, scale=0.75]
	
	\begin{axis}[
	height=5cm,		
	width=6cm,
	axis x line*=bottom, 
	axis y line*=left,
	ymin=80,
	grid=major,
	grid style={dashed,gray!50},
	ylabel = Accuracy (\%),
	xlabel = Number of includes per clause,
	legend style={at={(axis cs:8,88)},anchor=north,font=\footnotesize},
	]	
	\addplot [only marks, blue, opacity=0.3] table [y=acc, x=size] {Data/vanilla_tradeoff_gas.dat};
	\addlegendentry{Vanilla}
	\addplot [only marks, red, opacity=0.3] table [y=acc, x=size] {Data/diet_tradeoff_gas.dat};
	\addlegendentry{ETHEREAL}	
	\end{axis}
	
	\end{tikzpicture}
}
\subfloat[]{
	\begin{tikzpicture}[font=\normalsize, scale=0.75]
	
	\begin{axis}[
	height=5cm,		
	width=6cm,
	axis x line*=bottom, 
	axis y line*=left,
	ymin=70,
	grid=major,
	grid style={dashed,gray!50},
	ylabel = Accuracy (\%),
	xlabel = Number of includes per clause,
	legend style={at={(axis cs:25,75.5)},anchor=north,font=\footnotesize},
	]	
	\addplot [only marks, blue, opacity=0.3] table [y=acc, x=size] {Data/vanilla_tradeoff_gesture.dat};
	\addlegendentry{Vanilla}
	\addplot [only marks, red, opacity=0.3] table [y=acc, x=size] {Data/diet_tradeoff_gesture.dat};
	\addlegendentry{ETHEREAL}	
	\end{axis}
	
	\end{tikzpicture}
}
\subfloat[]{
	\begin{tikzpicture}[font=\normalsize, scale=0.75]
	
	\begin{axis}[
	height=5cm,		
	width=6cm,
	axis x line*=bottom, 
	axis y line*=left,
	grid=major,
	grid style={dashed,gray!50},
	ylabel = Accuracy (\%),
	xlabel = Number of includes per clause,
	legend style={at={(axis cs:60,85)},anchor=north,font=\footnotesize},
	]	
	\addplot [only marks, blue, opacity=0.3] table [y=acc, x=size] {Data/vanilla_tradeoff_har.dat};
	\addlegendentry{Vanilla}
	\addplot [only marks, red, opacity=0.3] table [y=acc, x=size] {Data/diet_tradeoff_har.dat};
	\addlegendentry{ETHEREAL}	
	\end{axis}
	
	\end{tikzpicture}
}
\vspace{-0.2cm}
\subfloat[]{
	\begin{tikzpicture}[font=\normalsize, scale=0.75]
	
	\begin{axis}[
	height=5cm,		
	width=6cm,
	axis x line*=bottom, 
	axis y line*=left,
	ymin=75,
	grid=major,
	grid style={dashed,gray!50},
	ylabel = Accuracy (\%),
	xlabel = Number of includes per clause,
	legend style={at={(axis cs:3,78.5)},anchor=north,font=\footnotesize},
	]	
	\addplot [only marks, blue, opacity=0.3] table [y=acc, x=size] {Data/vanilla_tradeoff_mammography.dat};
	\addlegendentry{Vanilla}
	\addplot [only marks, red, opacity=0.3] table [y=acc, x=size] {Data/diet_tradeoff_mammography.dat};
	\addlegendentry{ETHEREAL}	
	\end{axis}
	
	\end{tikzpicture}
}
\subfloat[]{
	\begin{tikzpicture}[font=\normalsize, scale=0.75]
	
	\begin{axis}[
	height=5cm,		
	width=6cm,
	axis x line*=bottom, 
	axis y line*=left,
	ymin=80,
	grid=major,
	grid style={dashed,gray!50},
	ylabel = Accuracy (\%),
	xlabel = Number of includes per clause,
	legend style={at={(axis cs:13,83.5)},anchor=north,font=\footnotesize},
	]	
	\addplot [only marks, blue, opacity=0.3] table [y=acc, x=size] {Data/vanilla_tradeoff_sensorless.dat};
	\addlegendentry{Vanilla}
	\addplot [only marks, red, opacity=0.3] table [y=acc, x=size] {Data/diet_tradeoff_sensorless.dat};
	\addlegendentry{ETHEREAL}	
	\end{axis}
	
	\end{tikzpicture}
}
\subfloat[]{
	\begin{tikzpicture}[font=\normalsize, scale=0.75]
	
	\begin{axis}[
	height=5cm,		
	width=6cm,
	axis x line*=bottom, 
	axis y line*=left,
	ymin=85,
	grid=major,
	grid style={dashed,gray!50},
	ylabel = Accuracy (\%),
	xlabel = Number of includes per clause,
	legend style={at={(axis cs:5.5,89)},anchor=north,font=\footnotesize},
	]	
	\addplot [only marks, blue, opacity=0.3] table [y=acc, x=size] {Data/vanilla_tradeoff_sports.dat};
	\addlegendentry{Vanilla}
	\addplot [only marks, red, opacity=0.3] table [y=acc, x=size] {Data/diet_tradeoff_sports.dat};
	\addlegendentry{ETHEREAL}	
	\end{axis}
	
	\end{tikzpicture}
}
\subfloat[]{
	\begin{tikzpicture}[font=\normalsize, scale=0.75]
	
	\begin{axis}[
	height=5cm,		
	width=6cm,
	axis x line*=bottom, 
	axis y line*=left,
	grid=major,
	grid style={dashed,gray!50},
	ylabel = Accuracy (\%),
	xlabel = Number of includes per clause,
	legend style={at={(axis cs:10,74.5)},anchor=north,font=\footnotesize},
	]	
	\addplot [only marks, blue, opacity=0.3] table [y=acc, x=size] {Data/vanilla_tradeoff_statlog.dat};
	\addlegendentry{Vanilla}
	\addplot [only marks, red, opacity=0.3] table [y=acc, x=size] {Data/diet_tradeoff_statlog.dat};
	\addlegendentry{ETHEREAL}	
	\end{axis}
	
	\end{tikzpicture}
}
\vspace{-0.2cm}
\caption{Trade-off between accuracy and model size, comparing vanilla and ETHEREAL TMs, for (a) EMG, (b) gas sensor, (c) GPS, (d) HAR, (e) mammographic mass, (f) sensorless drive diagnosis, (g) sport activity, and (h) statlog (vehicle silhouette).}
\label{fig:trade_off}

\Description[trade-off between accuracy and size]

\end{figure*}

We evaluate the trade-off between accuracy and model size for vanilla and ETHEREAL TMs, in Figure \ref{fig:trade_off}, based on metrics from each training epoch. The results show that ETHEREAL consistently offers a better trade-off by producing models with fewer includes while maintaining comparable accuracy. This is most notable in Figure \ref{fig:trade_off} (d), (e) and (h). Although ETHEREAL does not always reach the highest accuracy as the vanilla TM, it still exhibits a superior trade-off at lower accuracy levels.  

\vspace{-0.2cm}
\subsection{On-Platform Evaluation using STM32F746G-DISCO} \label{sec:stm32}
In the on-platform evaluation, both the vanilla and ETHEREAL TM are encoded using REDRESS and deployed on the micro-controller. ETHEREAL is expected to deliver shorter inference time, lower energy consumption, and a reduced memory footprint, compared to the REDRESS TM ($i.e.,$ the REDRESS-encoded vanilla TM).

\definecolor{new_red}{rgb}{0.95,0.47,0.47}
\definecolor{new_green}{rgb}{0.85,0.98,0.77}

\begin{table*}[!htb]
	\centering
	\caption{Comparisons between RFs, BNNs, REDRESS and ETHEREAL TMs on STM32F746G-DISCO micro-controller.}
    \vspace{-0.2cm}
	\label{tab:stm32_results}
\resizebox{\textwidth}{!}{%
	\begin{tabular}{|c||ccc||ccc||ccc||cccccc|}
	\hline
 	\multirow{2}{*}{} & \multicolumn{3}{c||}{\textbf{RF}} & \multicolumn{3}{c||}{\textbf{BNN: FC256} \cite{geiger2020larq}} & \multicolumn{3}{c||}{\textbf{REDRESS TM} \cite{maheshwari2023redress}} & \multicolumn{6}{c|}{\textbf{ ETHEREAL TM}} \\ \cline{2-16}
 	& \cellcolor{gray!66} {\begin{tabular}[c]{@{}c@{}} Time \\ (s) \end{tabular}} & \cellcolor{gray!33} {\begin{tabular}[c]{@{}c@{}} Mem \\ (kB) \end{tabular}} & {\begin{tabular}[c]{@{}c@{}} Energy \\ (mJ) \end{tabular}} & \cellcolor{gray!66} {\begin{tabular}[c]{@{}c@{}} Time \\ (s) \end{tabular}} & \cellcolor{gray!33} {\begin{tabular}[c]{@{}c@{}} Mem \\ (kB) \end{tabular}} & {\begin{tabular}[c]{@{}c@{}} Energy \\ (mJ) \end{tabular}} & \cellcolor{gray!66} {\begin{tabular}[c]{@{}c@{}} Time \\ (s) \end{tabular}} & \cellcolor{gray!33} {\begin{tabular}[c]{@{}c@{}} Mem \\ (kB) \end{tabular}} & {\begin{tabular}[c]{@{}c@{}} Energy \\ (mJ) \end{tabular}} & \cellcolor{gray!66} {\begin{tabular}[c]{@{}c@{}} Time \\ (s) \end{tabular}} & \cellcolor{gray!66} {\begin{tabular}[c]{@{}c@{}} Time \\ reduct. (\%) \end{tabular}} & \cellcolor{gray!33} {\begin{tabular}[c]{@{}c@{}} Mem \\ (kB) \end{tabular}} & \cellcolor{gray!33} {\begin{tabular}[c]{@{}c@{}} Mem \\ reduct. (\%) \end{tabular}} & {\begin{tabular}[c]{@{}c@{}} Energy \\ (mJ) \end{tabular}} & {\begin{tabular}[c]{@{}c@{}} Energy \\ reduct. (\%) \end{tabular}}\\ \hline 
 	EMG & \cellcolor{gray!66} 0.026 & \cellcolor{gray!33} 1394.80 & 6.13 & \cellcolor{gray!66} 1.51 & \cellcolor{gray!33} 643.36 & 355.97 & \cellcolor{gray!66} 0.72 & \cellcolor{gray!33} 1001.54 & 179.36 & \cellcolor{gray!66} 0.44 & \cellcolor{gray!66} 38.83 & \cellcolor{gray!33} 580.96 & \cellcolor{gray!33} 41.99 & 111.04 & 38.09 \\
 	Gas sensor & \cellcolor{gray!66} 0.027  & \cellcolor{gray!33} 1598.72 & 6.37 & \cellcolor{gray!66} 1.20 & \cellcolor{gray!33} 532.72 & 282.96 & \cellcolor{gray!66} 0.70 & \cellcolor{gray!33} 988.93 & 172.77 & \cellcolor{gray!66} 0.42 & \cellcolor{gray!66} 39.77 & \cellcolor{gray!33} 583.30 & \cellcolor{gray!33} 41.02 & 105.46 & 38.96 \\
 	GPS & \cellcolor{gray!66} 0.026 & \cellcolor{gray!33} 1531.54 & 6.09 & \cellcolor{gray!66} 1.66 & \cellcolor{gray!33} 504.24 & 388.61 & \cellcolor{gray!66} 2.40 & \cellcolor{gray!33} 3286.53 & 578.46 & \cellcolor{gray!66} 1.41 & \cellcolor{gray!66} 41.27 & \cellcolor{gray!33} 1933.25 & \cellcolor{gray!33} 41.18 & 341.25 & 41.01 \\
 	HAR & \cellcolor{gray!66} 0.024 & \cellcolor{gray!33} 1817.38 & 5.95 & \cellcolor{gray!66} 6.34 & \cellcolor{gray!33} 1944.56 & 1570.76 & \cellcolor{gray!66} 3.33 & \cellcolor{gray!33} 3613.74 & 810.47 & \cellcolor{gray!66} 0.29 & \cellcolor{gray!66} 91.26 & \cellcolor{gray!33} 512.27 & \cellcolor{gray!33} 85.82 & 121.25 & 85.04 \\
	Mammographic mass & \cellcolor{gray!66} 0.014 & \cellcolor{gray!33} 113.50 & 3.29 & \cellcolor{gray!66} 0.16 & \cellcolor{gray!33} 44.26 & 37.58 & \cellcolor{gray!66} 0.012 & \cellcolor{gray!33} 23.33 & 3.63 & \cellcolor{gray!66} 0.008 & \cellcolor{gray!66} 34.25 & \cellcolor{gray!33} 16.91 & \cellcolor{gray!33} 27.50 & 2.75 & 24.26 \\
	Sensorless drive & \cellcolor{gray!66} 0.031 &  \cellcolor{gray!33} 1928.29 & 7.78 & \cellcolor{gray!66} 1.41 & \cellcolor{gray!33} 2202.72 & 353.84 & \cellcolor{gray!66} 1.59 & \cellcolor{gray!33} 2210.90 & 400.74 & \cellcolor{gray!66} 0.91 & \cellcolor{gray!66} 42.39 & \cellcolor{gray!33} 1214.14 & \cellcolor{gray!33} 45.08 & 238.40 &  40.51 \\
	Sport activity & \cellcolor{gray!66} 0.025 & \cellcolor{gray!33} 1729.68 & 6.48 & \cellcolor{gray!66} 0.67 & \cellcolor{gray!33} 1467.60 & 173.71 & \cellcolor{gray!66} 0.54 & \cellcolor{gray!33} 776.64 & 144.75 & \cellcolor{gray!66} 0.36 & \cellcolor{gray!66} 33.52 & \cellcolor{gray!33} 459.81 & \cellcolor{gray!33} 40.80 & 100.65 & 30.46 \\
	Statlog & \cellcolor{gray!66} 0.024 & \cellcolor{gray!33} 1439.23 & 5.52 & \cellcolor{gray!66} 3.59 & \cellcolor{gray!33} 194.74 & 825.99 & \cellcolor{gray!66} 0.221 & \cellcolor{gray!33} 286.42 & 53.85 & \cellcolor{gray!66} 0.219 & \cellcolor{gray!66} 0.75 & \cellcolor{gray!33} 283.92 & \cellcolor{gray!33} 0.87 & 53.41 & 0.82 \\
 	\hline
	\end{tabular}
}
\vspace{-0.3cm}
\end{table*} 

Table \ref{tab:stm32_results} presents the inference time, memory footprint and energy obtained from the micro-controller for the RF, FC256 BNN, REDRESS TM, and ETHEREAL TM. The inference time and energy are averaged per datapoint, with energy measured using a Keithley DC power supply.
{Generally, RF provides the fastest inference and lowest energy consumption compared to BNN and TM, but it has the highest memory footprint due to its use of floating-point data representation. In contrast, BNN and TM result in more compact models by primarily using binary or Boolean values for logic operations, where ETHEREAL TM offers a 7$\times$ around reduction in memory footprint compared to RF, specifically for the case of mammography mass,}
This efficiency allows most ETHEREAL TM models to be deployed on state-of-the-art micro-controllers with up to 512 kB SRAM and 2 MB Flash \cite{lin2020mcunet}, unlike many other models in the comparison that exceed these limits.
Comparing the BNN and REDRESS TM across all datasets, both models generally exhibit similar memory footprints. Notably, TMs demonstrate significantly shorter inference time than BNNs for most datasets, with TMs achieving over 10$\times$ faster inference for mammographic mass and statlog.
This reduction in inference time also results in more than 10$\times$ lower energy consumption.
This is due to the high sparsity of a TA array, where most features are excluded from a TM after training, as described in Section \ref{sec:intro}. Consequently, these features are not used during inference, significantly reducing inference time and energy. In contrast, a BNN must consider all features during its inference process. 
While reducing the number of neurons or layers in a BNN could lower inference time and energy, it would also lead to a further decline in accuracy. Notably, the FC256 BNN has already demonstrated lower accuracy compared to the vanilla TMs, as shown in Table \ref{tab:acc_size}.
Finally, compared to REDRESS TMs, ETHEREAL TMs demonstrate reductions across all metrics, corresponding with the percentage decrease in the number of includes presented in Table \ref{tab:acc_size}.
This can be expected as ETHEREAL utilizes fewer includes during inference, which reduces both inference time and energy consumption, while also decreasing runtime memory usage. Furthermore, since REDRESS retains only the information of included literals, ETHEREAL further reduces the memory required to store the model.

\section{Conclusion} \label{sec:conc}
We introduced ETHEREAL, a model compression method for TM. ETHEREAL excludes insignificant literals based on their occurrences in both positive and negative clauses. 
This exclusion is facilitated by a modified training regime. Compared to the vanilla TM, ETHEREAL TM achieves up to an 87.54\% reduction in number of includes, while resulting in only a 3.38\% decrease in accuracy across eight TinyML applications. The reduction in model size leads to proportional reductions in inference time, memory footprint and energy, for micron-controller based implementations. 
{The reduction of accuracy is a reasonable compromise for substantial gains in inference speed and reduced resource consumption.} 
Compared to BNNs, ETHEREAL TM offers over 10$\times$ less inference time and energy; compared to RF, ETHEREAL TM provides up to 7$\times$ less memory usage. In summary, ETHEREAL enhances the trade-off between accuracy and model size, promoting efficient TM implementations with low cost, high speed and trustworthy behavior.


\begin{acks}
This work was supported by the Engineering and Physical Sciences Research Council (EPSRC) under Grant EP/X039943/1 and Grant EP/X036006/1.
\end{acks}

\bibliographystyle{ACM-Reference-Format}
\bibliography{acmart}

\end{document}